\documentclass[11pt]{article}

\usepackage[final]{acl}

\usepackage{times}
\usepackage{latexsym}

\usepackage[T1]{fontenc}
\usepackage{needspace}
\usepackage[utf8]{inputenc}

\usepackage{microtype}

\usepackage{inconsolata}

\usepackage{graphicx}

\usepackage{soul}
\usepackage{url}
\usepackage[utf8]{inputenc}
\usepackage{graphicx}
\usepackage{amsmath}
\usepackage{amsthm}
\usepackage{booktabs}
\usepackage{algorithm}
\usepackage{algorithmic}
\usepackage[switch]{lineno}
\usepackage{amssymb}
\usepackage{pifont}
\usepackage{xcolor}
\usepackage{amssymb}
\usepackage{colortbl}
\usepackage[normalem]{ulem}
\useunder{\uline}{\ul}{}
\usepackage[inline]{enumitem}
\usepackage{enumitem}
\definecolor{customgreen}{rgb}{0.0, 0.5, 0.0} 
\usepackage{arydshln}
\usepackage{tikz}
\definecolor{table_yellow}{RGB}{254,243,222}
\definecolor{table_blue}{RGB}{237,245,255}
\definecolor{table_green}{RGB}{234,245,234}
\usepackage{multirow}
\definecolor{caption_green}{RGB}{112,173,71}
\definecolor{caption_blue}{RGB}{60,120,216}
\definecolor{caption_pink}{RGB}{255,175,204}

\title{PLANE: Plug-and-Play Fine-Grained Visual Experts for Hallucination Mitigation in Large Vision-Language Models}

\author{
\bf{
Jianghao Yin$^{1}$, Lang Yu$^{1}$, Qianxi Yu$^1$, Qin Chen$^1$\thanks{Corresponding Author.}, Jie Zhou$^1$,}\\
\bf{
 Zhikai Lei$^2$, Liang He$^1$}\\
{$^1$School of Computer Science and Technology, East China Normal University}\ \\ 
{$^2$Shanghai Qiji Zhifeng Co., Ltd.}\vspace{2mm}\\
\texttt{jhyin@stu.ecnu.edu.cn, qchen@cs.ecnu.edu.cn} 
}

\begin{document}
\maketitle
\begin{abstract}
Large Vision-Language Models (LVLMs) commonly suffer from hallucinations, which appear as fabricating non-existent objects or outputting incorrect attributes and spatial relationships in images. Previous works empirically indicate that CLIP-style vision encoders tend to focus on primary objects while ignoring image details, which is one of the causes of hallucinations. To address this issue, we propose plug-and-play fine-grained visual experts (PLANE) comprising two components: LoRA visual experts operating on CLIP for multi-aspect image encoding, and an instruction-aware gate network for adaptive image feature fusion.  During training, we construct a fine-grained image-caption pairs dataset, where images are described from three aspects: objects, attributes, and relations between objects. Image-caption pairs from different aspects are used to train the three fine-grained LoRA experts respectively.  Aiming at 
comprehensively evaluate our method's capabilities, we manually annotate a fine-grained hallucination evaluation benchmark for LVLMs which is GPT4-independent during evaluation phase (GIFT). We conduct experiments both on popular hallucination evaluation benchmarks and GIFT. The results show that our method effectively reduces hallucinations across models of different types and sizes while introducing minimal inference latency. Compared to advanced baselines, our model also demonstrates superior performance.
\end{abstract}

\section{Introduction}

\label{sec:intro}
\begin{figure}[t]
    \begin{center}
        \includegraphics[width=1.0\columnwidth]{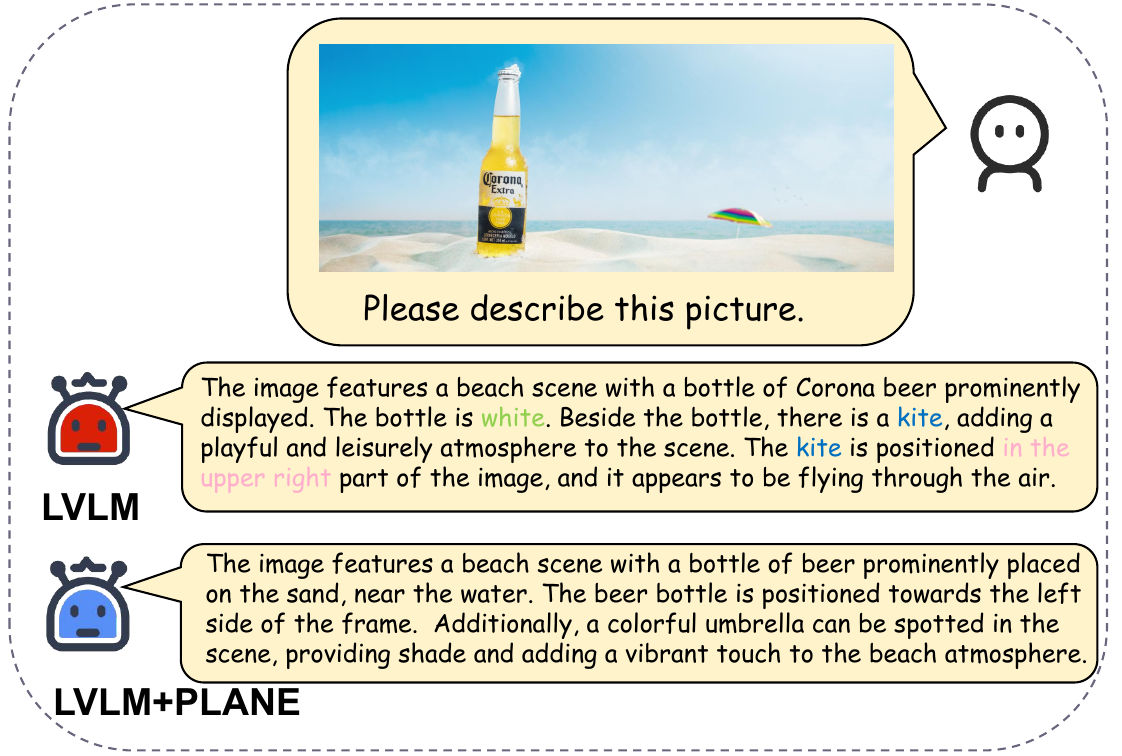}
    \end{center}
    \caption{A hallucination case in LVLMs, with different colors representing object hallucinations, attribute hallucinations, and relation hallucinations.}
    \label{fig:intro}
\end{figure}

Large Language Models (LLMs) have demonstrated exceptional generation and reasoning capabilities \cite{gpt4,llama3,qwen2,mixtral,baichuan,chatglm}. Building on this foundation, Large Vision-Language Models (LVLMs) have rapidly developed, typically consisting of a CLIP-style vision encoder \cite{clip}, a projection layer, and an LLM \cite{MMsurvey1,MMsurvey2,llavav1.5,qwenvl,instructblip,otter,varytoy}. They have shown excellent performance in tasks such as image captioning \cite{caption,blip2} and visual question answering (VQA) \cite{vqa,vqa2}. However, similar to LLMs \cite{LLMhallu1, LLMhallu2}, LVLMs also suffer from severe hallucination issues \cite{lvhallu,mmhallucination,vcd,woodpecker,lessismore,helpd,Causal,dmas,multimelo}. As illustrated in Figure \ref{fig:intro}, previous work categorizes multimodal hallucinations into three types: object hallucinations, attribute hallucinations, and relation hallucinations \cite{lvhallu}. These hallucinations undermine the credibility of LVLMs and hinder their application in various fields.

Many methods have been proposed to mitigate hallucinations in LVLMs, and these can generally fall into three types. The first type is decoding-based methods, such as VCD \cite{vcd} and OPERA \cite{opera}, which introduce specific strategies during the inference stage to influence the model's final output. While these methods typically do not require additional training, they will increase the inference time. The second type involves post-hoc methods, such as Volcano \cite{volcano} and LURE \cite{LURE} which use a revisor model to correct hallucinations in the generated text. However, these approaches often increase the time for obtaining the final output and are unable to address the inner cause of hallucinations within the model. The third type is fine-tuning-based methods, such as LRV \cite{LVR}. These methods generally create more robust datasets to fine-tune the LVLMs. However, a drawback of this approach is that it usually requires retraining when improving different LVLMs, consuming significant computational resources. Given the shortcomings of existing multimodal hallucination mitigation methods, can we mitigate hallucinations from the perspective of the internal causes that trigger hallucinations in LVLMs and propose an efficient plug-and-play solution that can be directly integrated into multiple LVLMs without requiring retraining?

Previous work has pointed out that CLIP-style vision encoders \cite{clip} tend to focus on primary objects while neglecting the details of images, which can also lead to hallucinations in LVLMs \cite{vcoder,cho2022fine,lvhallu}.  Based on these insights, this work aims to address the hallucination issue through enhancement of the vision encoder in LVLMs. An intuitive idea is to train a vision encoder by aligning it with highly detailed captions. However, the CLIP text encoder can only encode 77 tokens \cite{denseimage}, which significantly limits the level of detail that captions can provide about the image. Therefore, we decompose a detailed caption into multiple aspects to train several expert encoders.

Based on the above analysis, we propose \underline{\textbf{p}}lug-and-p\underline{\textbf{l}}ay fine-gr\underline{\textbf{a}}i\underline{\textbf{n}}ed visual \underline{\textbf{e}}xperts (PLANE) which consists of three fine-grained LoRA \cite{lora} visual experts operating on CLIP and an instruction-aware gate network. The former encodes images at object-aspect, attribute-aspect, and relation-aspect, while the latter fuses different features at the token-level according to instructions. The fused features are then used for LVLMs inference. To train the model, we constructed a fine-grained image-caption pairs dataset, which consists of three types of image-caption pairs: image-object caption pairs, image-attribute caption pairs, and image-relation caption pairs. The object captions primarily describe the objects present in the image, the attribute captions describe various attributes of the objects in the image, such as color, texture, shape, size, etc., and the relation captions describe the spatial relationships between objects in the image. Then, we use the constructed three types image-caption pairs to train three different LoRA visual expert, with each expert specializing in capturing different aspects of the image features. 

Currently, existing multimodal generative hallucination evaluation benchmarks have limitations: some only focus on object-level hallucination \cite{chair, LVR, Amber}, while others, although considering fine-grained hallucination, rely on ChatGPT or GPT-4 for assessment \cite{faithscore}, making the evaluation costly and difficult to scale. To comprehensively evaluate model performance, we have constructed a \underline{\textbf{G}}PT4-\underline{\textbf{i}}ndependent (during evaluation phase) \underline{\textbf{f}}ine-grained hallucina\underline{\textbf{t}}ion evaluation benchmark (GIFT) which is manually annotated. This benchmark enables the evaluation of LVLMs' hallucinations at the object-level, attribute-level, and relation-level in open-ended generation tasks.
In summary, the main contributions of our paper are:
\begin{itemize}

\item We propose \underline{\textbf{p}}lug-and-p\underline{\textbf{l}}ay fine-gr\underline{\textbf{a}}i\underline{\textbf{n}}ed visual \underline{\textbf{e}}xperts (PLANE) for hallucination mitigation, which is trained on our constructed fine-grained image-caption pairs dataset covering object-aspect, attribute-aspect, and relation-aspect. Once trained, PLANE can be directly integrated into other LVLMs that utilize the same vision encoder without requiring retraining.

\item We construct a \underline{\textbf{G}}PT4-\underline{\textbf{i}}ndependent (during evaluation phase) \underline{\textbf{f}}ine-grained hallucina\underline{\textbf{t}}ion evaluation benchmark (GIFT) that differs from previous object-level focused generative benchmarks. Our benchmark evaluates model hallucinations at object, attribute, and relation levels.

\item We conduct detailed experiments on different task benchmarks. The results show that our method effectively reduces hallucinations across models of different types and sizes while introducing minimal inference latency. Compared to advanced baselines, our model also demonstrates superior performance.
\end{itemize}

\section{Related Work}
\label{sec:formatting}

\subsection{Hallucinations in LVLMs}

To mitigate hallucinations in LVLMs, numerous methods have been proposed. VCD \cite{vcd} and OPERA \cite{opera} modify decoding strategies but increase inference time. Volcano \cite{volcano} and Woodpecker \cite{woodpecker} use post-hoc correction with additional models, which doesn't address the root cause and also increases inference time. LRV \cite{LVR} constructs more robust training datasets but requires retraining for different models. Recently, some work has focused on image encoders. Liu et al. \cite{OHD} explored object hallucinations in CLIP, using fine-grained object-level contrastive loss to fine-tune the CLIP model and reduce hallucinations in LVLMs. However, they only focus object-level hallucinations. 
Our goal is to propose a vision encoder optimization method that can mitigate hallucinations at multiple levels.

\subsection{Hallucination Benchmarks for LVLMs}
\label{sec:benchmarks}

Existing LVLM hallucination evaluation benchmarks assess hallucination through discriminative and generative tasks \cite{lvhallu, mmhallucination}. Discriminative tasks like POPE \cite{pope} and MME \cite{MME} require binary Yes/No answers, with POPE focusing on object-level hallucination and MME covering objects, attributes, and relations. Generative tasks evaluate hallucinations in image descriptions using benchmarks like GAVIE \cite{LVR} and Faithscore \cite{faithscore}, which rely on GPT-4 or ChatGPT as judges. While enabling fine-grained multi-level evaluation, their dependence on GPT-4/ChatGPT limits scalability. Alternative approaches like CHAIR \cite{chair} and AMBER \cite{Amber} avoid GPT-4 but are restricted to object-level assessment. To address this limitation, we propose GIFT, a GPT-4-independent benchmark (during evaluation phase) for comprehensive fine-grained hallucination evaluation.

\section{Method}

\begin{figure}[t]
    \begin{center}
        \includegraphics[width=1.0\columnwidth]{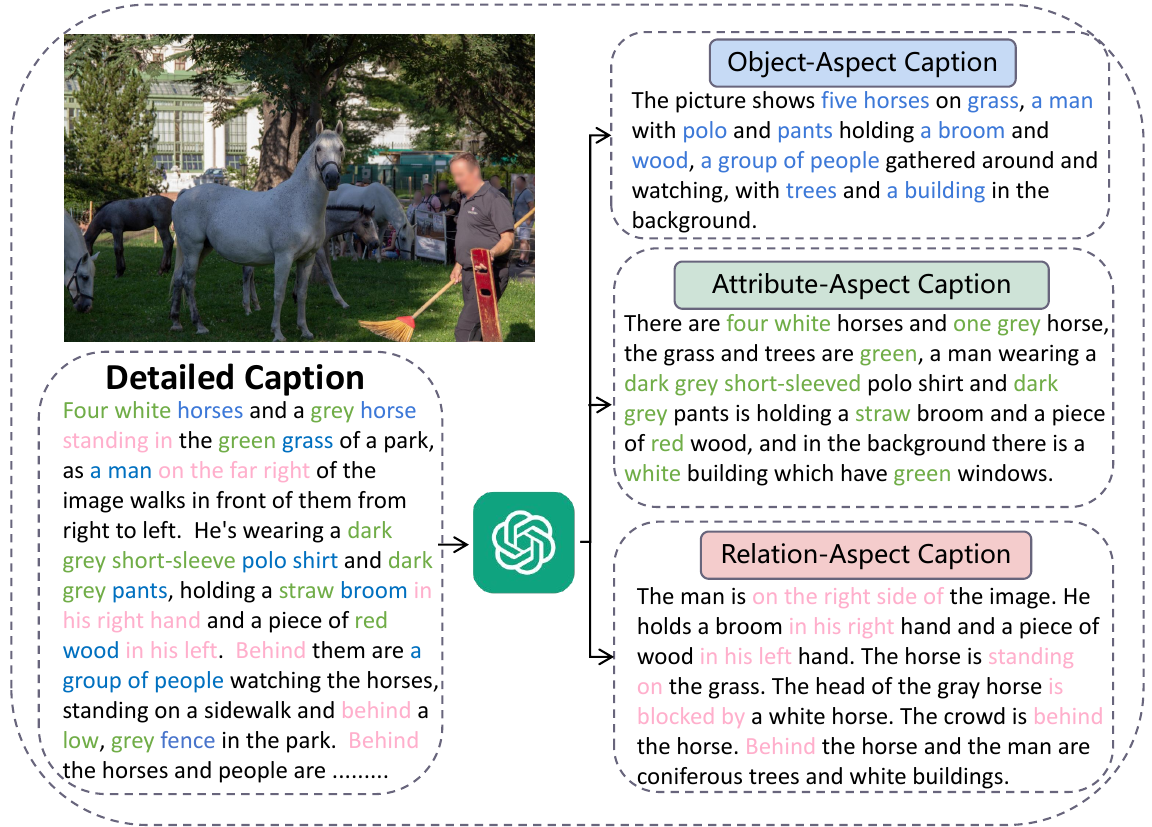}
    \end{center}
    \caption{The construction process for the fine-grained image-caption pairs dataset, where different colors indicate descriptions related to object, attribute, and relation.}
    \label{fig:introb}
\end{figure}

\begin{figure*}[thb!]
  \centering
  \includegraphics[width=1\textwidth]{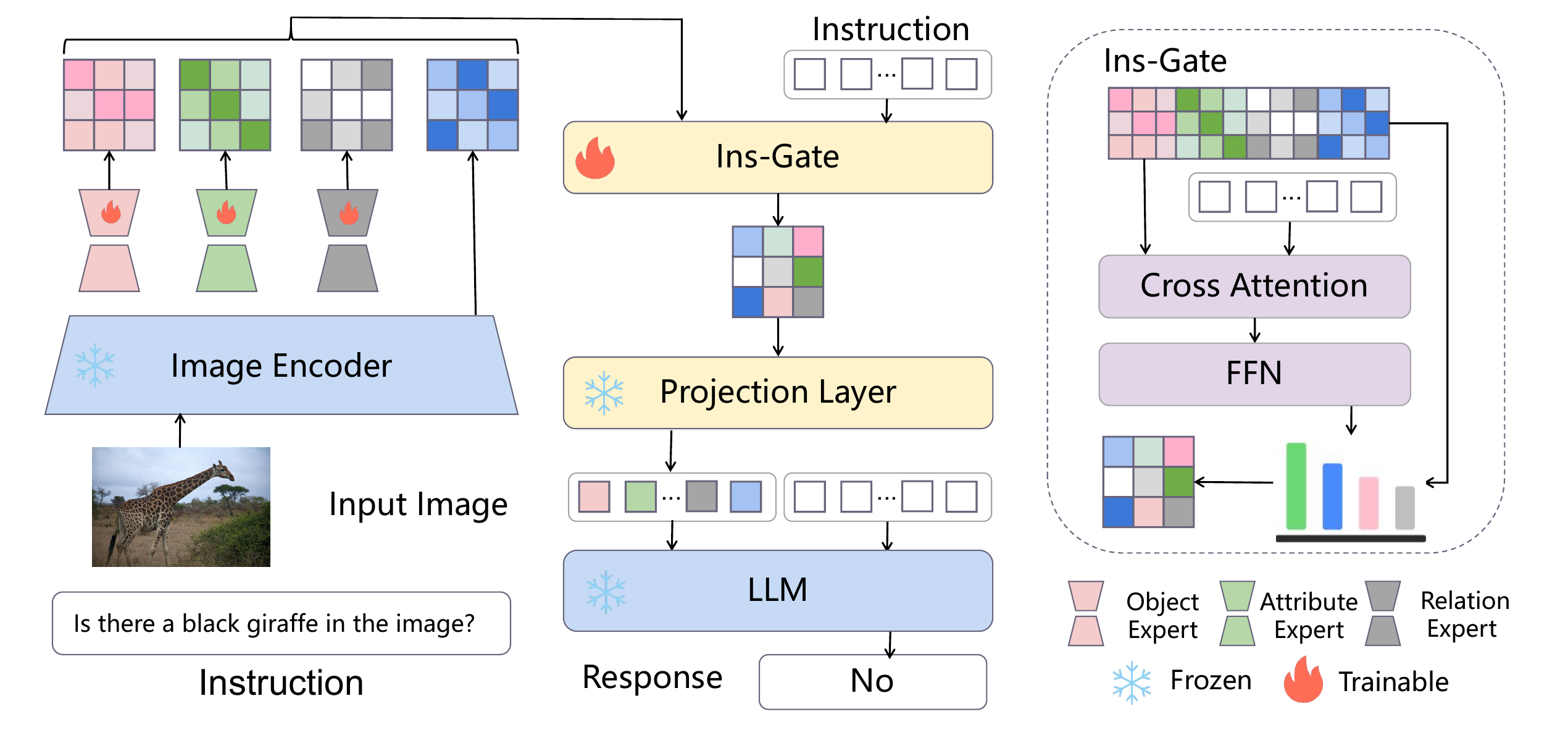}\\
  \caption{Overview of the PLANE. The input image is encoded by three LoRA visual experts and the original LVLM encoder, generating multi-level image features. These features are fused by Ins-Gate based on the instruction for LVLM inference.}
  \label{framework}
\end{figure*}

The model architecture of PLANE is demonstrated in Figure \ref{framework}. The main idea is to encode images with multiple visual experts that focus on different aspects, obtaining diverse visual features, which are then fused based on the instruction, and finally, the fused features are used for LVLMs inference. In this section, we will explain PLANE's training data construction, model architecture, and training methods.

\subsection{Dataset Construction}
\label{sec:dataset}
Urbanek et al. \cite{denseimage} develop the Densely Captioned Images (DCI) dataset, a manually annotated, dense and highly-aligned image-text pair dataset. They also demonstrate that fine-tuning vision-language models (VLMs) using a summarized version of this small dataset improved VLM performance. Inspired by this work, we build our fine-grained image-caption dataset based on the DCI dataset. As shown in Figure \ref{fig:introb}, we use ChatGPT to decompose the dense image captions into three categories: object captions, attribute captions, and relation captions. The object caption describes what is present in the image, the attribute caption focuses on the characteristics of each object, such as color, texture, and material, while the relation caption highlights the spatial relationships between objects in the image. We ensure that the length of each caption is kept within 77 tokens, and together these three captions aim to cover the content of the original dense caption as comprehensively as possible. For captions exceeding 77 tokens in length, ChatGPT will be called again to summarize them to within 77 tokens. To increase data diversity, we filter out a set of image-caption pairs related to attributes and position from the MSCOCO dataset \cite{coco} and incorporate them into our dataset. Finally, we construct a fine-grained image-caption pairs dataset with a total of 38,375 samples, which consists of 7,800 image-object caption pairs, 18,121 image-attribute caption pairs, and 12,454 image-relation caption pairs.

Previous studies show that hard negative captions during CLIP training can improve model performance \cite{negclip, denseimage}. Therefore, we design hallucination captions that maintain similar structure but differ semantically from original captions. Using ChatGPT, we generate three types of hallucinations for different fine-grained captions : object hallucinations by replacing or adding non-existent objects, attribute hallucinations by modifying descriptive terms (e.g., \textquotesingle wooden spoon\textquotesingle~to \textquotesingle glass spoon\textquotesingle), and relation hallucinations by altering spatial relationships (e.g., \textquotesingle birds on the tree\textquotesingle~to \textquotesingle birds under the tree\textquotesingle).

Formally, our dataset $D = \{D_i\},\quad i \in \{\textit{obj},\textit{attr}, \textit{rela}\}$, where $i$ indicates the specific aspect of the caption-image pairs, with $obj$, $attr$, and $rela$ respectively representing object, attribute, and relation. Each subset $D_i$ corresponds to a fine-grained dataset, which consists of multiple triplets $(I, C^i, C^{i-})$, where $I$ is the image, $C^i$ is the fine-grained caption, and $C^{i-}$ is the hallucination caption.

\subsection{Model Architecture}
Our model consists of two parts: fine-grained visual experts and an instruction-aware gate network (Ins-Gate). The former will apply to LVLMs vision encoder and capture different aspects of image representations, while the latter fuses these aspects into a new representation based on instruction.

\paragraph{\textbf{Fine-grained LoRA visual experts.}} These three LoRA visual experts are trained on object caption-image pairs, attribute caption-image pairs, and relation caption-image pairs, which are constructed in last section.

Given an input \((I, T)\), with image \(I\) and instruction \(T\). The image \(I\) is encoded by the original vision encoder from LVLMs into \(z_{ori} \in \mathbb{R}^{n\times d}\). At the same time, the image will also be encoded by the other three expert encoders, as shown in Equation \ref{eq:encoder}:
\begin{equation}
z_i = E_{\text{i}}(I), \quad i \in \{\textit{obj}, \textit{attr}, \textit{rela}\}
\label{eq:encoder}
\end{equation}
where \(E_i\) represents the vision encoder after applying the respective LoRA expert to the original encoder, and \(z_{i} \in \mathbb{R}^{n\times d}\) is the associated fine-grained visual embedding.

\paragraph{\textbf{Ins-Gate.}} Ins-Gate contains a cross-attention layer and a feed-forward network (FFN) layer. The instruction embedding \(t\), which is encoded from instruction \(T\), and the concatenated features \(z_c\) (representations from three aspects and the original encoder's representations) are used as inputs. The instruction embeddings and image features interact fully in the cross-attention layer, extracting image features relevant to the user's instruction from the four types of image representations. Then \(gate \in \mathbb{R}^{n\times 4}\) is computed using Equation~\ref{eq:gate}.
\begin{equation}
gate = Softmax(FFN(Attention(z_c,t)))
\label{eq:gate}
\end{equation}
The \(gate\) provides token-level fusion weights, where each row represents the weights assigned to each image token from the four types of image representations. The feature fusion process is described in Equation \ref{eq:fusion}:
\begin{equation}
f_{fusion} = \sum_{j=1}^{4} gate[:, j] \odot z_{\text{j}}
\label{eq:fusion}
\end{equation}
\(f_{fusion}\) represents the fused image features, which are concatenated with the text tokens to continue the forward process of the LVLMs.

\subsection{Two-Stage Training}
The training process is divided into two stages. Throughout the entire training process, we only train PLANE, while all original parameters of the LVLM were frozen. In the training stage \uppercase\expandafter{\romannumeral1}, different fine-grained LoRA visual experts are trained following the CLIP training paradigm \cite{clip}. In the training stage \uppercase\expandafter{\romannumeral2}, PLANE will be integrated into the LVLM, keeping only the Ins-Gate trainable, following the end-to-end LLaVA training paradigm \cite{llavav1.5}.
\paragraph{\textbf{Training Stage \uppercase\expandafter{\romannumeral1}}}
At this stage, the fine-grained image-caption pairs dataset $D$ is used to train the LoRA visual experts. Each of the three LoRA experts is trained on $D_{obj}$, $D_{attr}$, and $D_{rela}$, respectively, to enhance the image encoder's ability to capture aspect-specific features. Following prior work on VLMs \cite{clip,denseimage}, we employ contrastive learning loss $\mathcal{L}_{CL}$ to train fine-grained image encoders. For a batch of $N$ image-caption pairs from dataset $D_i$, $\mathcal{L}_{CL}$ combines bidirectional image-to-text and text-to-image contrastive losses using cosine similarity $sim$ and temperature parameter $\tau$.

Inspired by previous work \cite{negclip,denseimage}, we incorporate a negative loss $\mathcal{L}_{NL}$ to enhance the vision-language model's discriminative capability, where hallucination captions serve as carefully designed negative samples for each aspect:

\begin{equation}
\begin{aligned}
\mathcal{L}_{NL} &= - \frac{1}{N} \sum_{k=1}^{N} \log \bigg( \exp\big(sim(I_k^i, C_k^i)/\tau\big) \times \\
&\Big[ \sum_{C^{i*}\in\{C^i,C^{i-}\}} \exp\big(sim(I_k^i, C^{i*})/\tau\big) \Big]^{-1} \bigg)
\end{aligned}
\end{equation}
Here, $(I_k^i, C_k^i)$ represents an image-caption pair from $D_i$, and $C_k^{i-}$ is the corresponding hallucination caption. The final training loss combines both objectives:
\begin{equation}
\mathcal{L}_{final} = \alpha \mathcal{L}_{CL} + \beta \mathcal{L}_{NL}
\label{eq:finalloss}
\end{equation}
where $\alpha$ and $\beta$ are hyperparameters.

\paragraph{\textbf{Training Stage \uppercase\expandafter{\romannumeral2}}}
At this stage, only the Ins-Gate is trainable, while all other modules remain frozen. We use the LRV-instruction dataset \cite{LVR} to train the Ins-Gate with the autoregressive loss \cite{llavav1.5}.

\section{GIFT : GPT4-Independent Fine-Grained Hallucination Evaluation Benchmark}
\label{sec:GIFT}

\begin{figure}[t]
    \begin{center}
        \includegraphics[width=1.0\columnwidth]{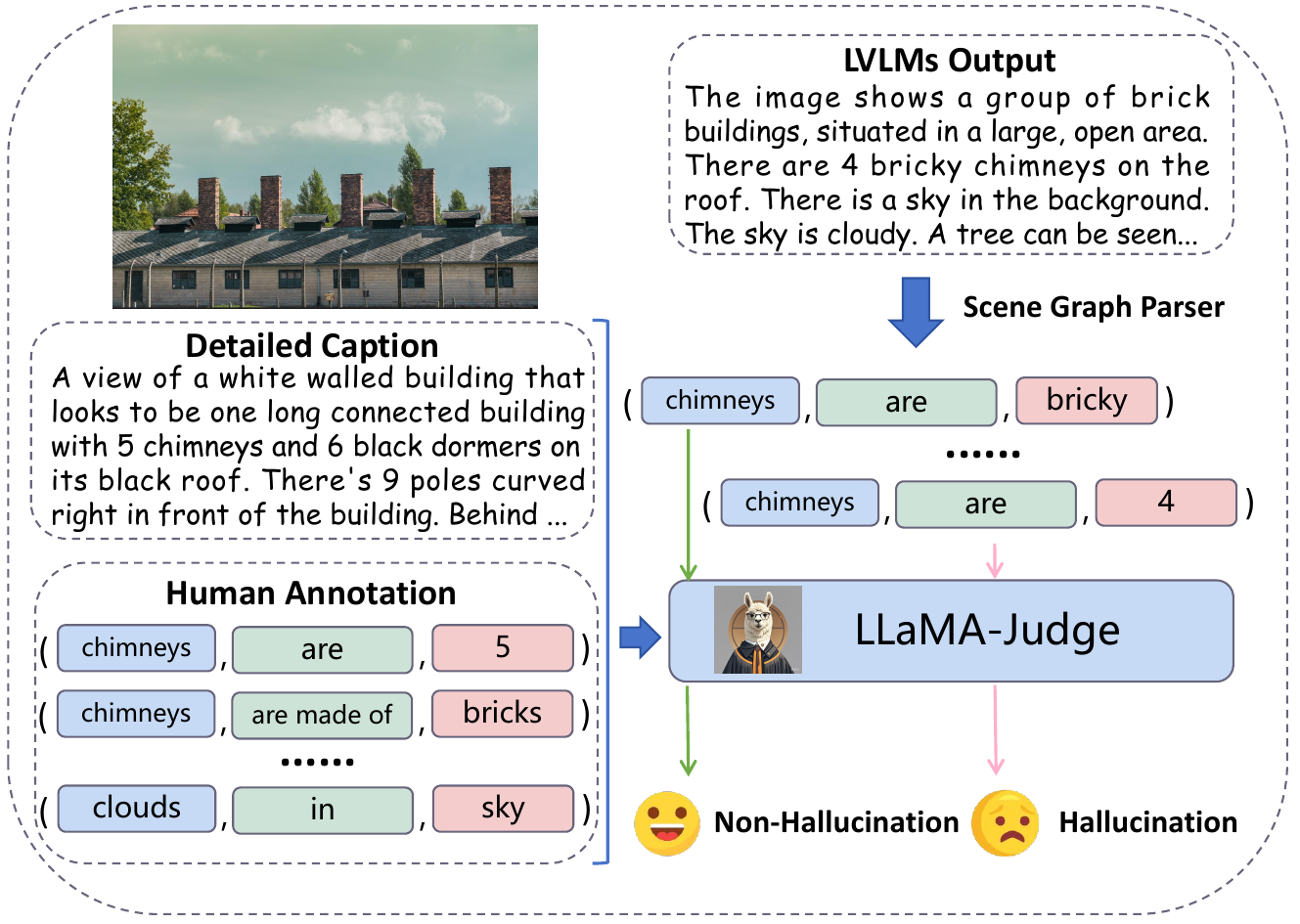}
    \end{center}
    \caption{Pipeline of GIFT. The LVLM output is fed into a scene graph parser to extract triplets, which are then input into Llama-Judge along with the detailed caption and manually annotated triplets for hallucination detection.}
    \label{fig:GIFT}
\end{figure}
As discussed in related work, there is a lack of fine-grained hallucination evaluation benchmarks for generation tasks that can function independently of GPT-4. Therefore, we propose a GPT4-independent fine-grained benchmark (during evaluation phase) to comprehensively assess our models.
We build our benchmark based on the DCI dataset \cite{denseimage}, which consists of manually annotated detailed text-image pairs. We select 206 images from its validation and test sets, and manually annotate 2,120 scene graph triplets (including object, attribute,and relation) aiming to cover as many image details as possible.

The evaluation pipeline of GIFT is shown in Figure \ref{fig:GIFT}, LVLMs are first prompted with \textquotesingle Please describe the image in as much detail as possible\textquotesingle~to generate captions for each image. Then, scene graph parser \cite{parser} is used to extract triplets contained in the captions. To determine whether these triples are hallucinating compared to manually annotated triples, we train a model called Llama-Judge based on Llama 3 \cite{llama3}. These hallucinations may include fabricating non-existent objects, incorrect attributes, or relationships, which help determine if hallucination occurs. We introduce two metrics, Hallu Rate and Cover Rate, to measure the extent of hallucination and level of detail in model-generated content. The definition of Hallu Rate is shown in Equation \ref{eq:hallu}:
\begin{equation}
\text{Hallu Rate} = \frac{H}{M} \times 100\% 
\label{eq:hallu}
\end{equation}
$H$ represents the number of hallucinated triplets detected by Llama-Judge, $M$ represents the total number of triplets extracted from LVLMs' generation. The definition of Cover Rate is shown in Equation \ref{eq:cover}:
\begin{equation}
\text{Cover Rate} = \frac{NH}{K} \times 100\% 
\label{eq:cover}
\end{equation}
$NH$ represents the number of triplets judged as non-hallucinated, and $K$ represents the number of manually annotated triplets. More details about GITF are presented in the Appendix \ref{app:2}.

\section{Experimental Setup}

\subsection{Datasets and Implementation Details}

We evaluate our model on discriminative tasks and open-ended generation tasks. For discriminative tasks, we select \textbf{MME} \cite{MME} and \textbf{POPE} \cite{pope} (the experimental results for POPE are provided in the Appendix \ref{app:pope}). For open-ended generation tasks, we select \textbf{AMBER} \cite{Amber} and \textbf{GIFT}. More details about the benchmarks and metrics are provided in the Appendix \ref{app:benchmark}.

\subsection{Baselines and Implementation Details}
We evaluate PLANE on several LVLMs of different sizes: LLaVA-v1.5 7B \cite{llavav1.5}, Mini-Gemini 7B \cite{mini-gemini}, Parrot 7B \cite{parrot}, LLaVA 13B \cite{llava}, Otter 7B \cite{otter}, and Vary Toy 1.8B \cite{varytoy}. Our model is trained on LLaVA-v1.5 7B, and after training, it is directly applied to other LVLMs in a plug-and-play manner.  Furthermore, we conduct comprehensive comparisons between our proposed method and current advanced approaches in both categories: training-free models and those requiring fine-tuning. Among training-free methods, we benchmark against VCD \cite{vcd}, DROPOUT DECODING \cite{dropout} and VAF \cite{VAF}. For approaches requiring training, we compare with OHD-Caps \cite{OHD}, HADPO \cite{hadpo}, and HALVA \cite{halva} to evaluate our method's effectiveness. All experiments are conducted on two NVIDIA A800(80GB) GPUs.

In training stage \uppercase\expandafter{\romannumeral1}, we train visual experts on two vision encoders CLIP-ViT-L-224 (which is used for Otter, Vary Toy, LLaVA) and CLIP-ViT-L-336 (which is  used for LLaVA-1.5, Mini-Gemini, Parrot). We set the LoRA rank to 32, the batch size to 32, and both alpha and beta to 1, running for up to 10 epochs. In training stage \uppercase\expandafter{\romannumeral2}, we integrate our approach into LLaVA-1.5 to train the Ins-Gate. The batch size is 128, running for up to 2 epochs. For evaluation, we set the temperature to 0.5 for all methods. For a more convincing comparison, we reproduce all the baselines based on the officially released code and run each experiment  over 3 times, reporting the average results.

 \section{Experiment}
\label{sec:Experiment}

\begin{table*}[thb!]
\setlength{\tabcolsep}{1pt}
  \centering
 \resizebox{\textwidth}{!}{
  \begin{tabular}{ccccccc}
    \toprule
    Model&  Method & Existence$\uparrow$ & Count $\uparrow$& Position$\uparrow$& Color$\uparrow$ & Total$\uparrow$  \\
    \midrule
\multirow{9}{*}{LLaVA-1.5 7B}  & Original & $179.00_{(\pm2.24)}$ & $115.00_{(\pm15.76)}$ & $104.67_{(\pm20.26)}$ & $156.71_{(\pm4.88)}$ & 555.38 \\
                             & OHD-Caps & $180.00_{(\pm3.54)}$ & $120.67_{(\pm9.47)}$ & $101.67_{(\pm17.36)}$ & $157.00_{(\pm9.08)}$ & 559.34 \\
                             & HA-DPO & $183.33_{(\pm2.89)}$ & $118.89_{(\pm7.88)}$ & $112.78_{(\pm3.47)}$ & $149.44_{(\pm5.85)}$ & 564.44 \\
                               & HALVA & $181.67_{(\pm2.89)}$ & $116.33_{(\pm14.43)}$ & $\mathbf{122.33_{(\pm1.92)}}$ & $160.00_{(\pm10.00)}$ & 580.33 \\
                               & VCD & $182.00_{(\pm2.74)}$  & $128.00_{(\pm7.58)}$ & $115.00_{(\pm15.76)}$ & $147.66_{(\pm20.26)}$ & 572.66 \\
                                & DROPOUT DEC. & $183.00_{(\pm5.70)}$  & $121.33_{(\pm12.38)}$ & $119.67_{(\pm3.41)}$ & $148.67_{(\pm6.61)}$ & 572.67 \\
                              & VAF & $183.33_{(\pm5.77)}$  & $\mathbf{128.33_{(\pm7.58)}}$ & $120.56_{(\pm3.47)}$ & $149.44_{(\pm5.09)}$ & 581.66 \\
                                & PLANE & $\mathbf{184.00_{(\pm4.18)}}$  & $119.67_{(\pm5.05)}$ & $112.00_{(\pm13.10)}$ & $\mathbf{166.67_{(\pm12.19)}}$ & \textbf{582.34} \\
                             & $\Delta$ & \textbf{$\uparrow${5.00}} & \textbf{$\uparrow${4.67}} & \textbf{$\uparrow${7.33}} & \textbf{$\uparrow${9.96}} & \textbf{$\uparrow${26.96}} \\

\midrule
\multirow{4}{*}{Mini-Gemini 7B} & Original &  $179.44_{(\pm11.34)}$ & $132.22_{(\pm11.09)}$ & $109.44_{(\pm7.87)}$ & $139.44_{(\pm9.62)}$ & 560.54 \\
                              & VCD &  $167.77_{(\pm11.09)}$ & $126.11_{(\pm23.11)}$ & $105.00_{(\pm2.89)}$ & $141.67_{(\pm17.64)}$ & 540.55 \\
                               & PLANE & $\mathbf{180.00_{(\pm5.00)}}$ & $\mathbf{137.22_{(\pm11.09)}}$ & $\mathbf{119.45_{(\pm13.47)}}$ &$\mathbf{149.44_{(\pm5.85)}}$ & \textbf{586.11} \\
                               & $\Delta$ & \textbf{$\uparrow${0.56}} & \textbf{$\uparrow${5.00}} & \textbf{$\uparrow${10.01}} & \textbf{$\uparrow${10.00}} & \textbf{$\uparrow${25.57}} \\

\midrule
\multirow{4}{*}{Parrot 7B} & Original &  $183.33_{(\pm2.89)}$ & $138.89_{(\pm2.55)}$ & $137.78_{(\pm0.96)}$ & $166.11_{(\pm7.88)}$ & 626.12 \\
                              & VCD &  $\mathbf{185.00_{(\pm0.00)}}$ & $143.67_{(\pm17.04)}$ & $133.30_{(\pm8.72)}$ & $168.33_{(\pm5.77)}$ & 630.30 \\
                               & PLANE & $\mathbf{185.00_{(\pm5.00)}}$ & $\mathbf{144.45_{(\pm9.48)}}$ & $\mathbf{145.00_{(\pm2.89)}}$ &$\mathbf{173.33_{(\pm2.89)}}$ & \textbf{647.78} \\

                            & $\Delta$ & \textbf{$\uparrow${1.67}} & \textbf{$\uparrow${5.56}} & \textbf{$\uparrow${7.22}} & \textbf{$\uparrow${7.22}} & \textbf{$\uparrow${21.66}} \\

\midrule
\multirow{4}{*}{Otter 7B} & Original & $163.00_{(\pm4.47)}$ & $102.67_{(\pm7.51)}$ & $66.67_{(\pm7.07)}$ & $113.33_{(\pm3.54)}$ & 445.67 \\
                                & VCD & $170.00_{(\pm3.54)}$ & $96.33_{(\pm10.89)}$ & $67.33_{(\pm6.73)}$ & $114.67_{(\pm8.11)}$ & 448.33 \\
                              & PLANE & $\mathbf{174.00_{(\pm2.23)}}$ & $\mathbf{105.00_{(\pm2.63)}}$ & $\mathbf{70.67_{(\pm3.65)}}$ & $\mathbf{115.00_{(\pm4.57)}}$ & \textbf{464.67} \\
                               &  $\Delta$ &  \textbf{$\uparrow${11.00}} &  \textbf{$\uparrow${2.33}} &  \textbf{$\uparrow${4.00}} &  \textbf{$\uparrow${1.67} }&  \textbf{$\uparrow${19.00}} \\

\midrule                              
\multirow{4}{*}{Vary Toy 1.8B}  & Original & $176.33_{(\pm2.74)}$ & $86.00_{(\pm10.31)}$ & $84.33_{(\pm9.76)}$ & $87.00_{(\pm2.74)}$  & 433.66 \\
                                & VCD & $176.67_{(\pm5.00)}$ & $93.34_{(\pm6.24)}$ & $85.67_{(\pm8.04)}$ & $94.00_{(\pm4.18)}$  & 449.68 \\
                               & PLANE & $\mathbf{183.67_{(\pm3.20)}}$ & $\mathbf{99.00_{(\pm6.93)}}$ & $\mathbf{85.67_{(\pm1.90)}}$ & $\mathbf{103.00_{(\pm4.47)}}$  & \textbf{471.34} \\
                               & $\Delta$ & \textbf{$\uparrow${7.34}} & \textbf{$\uparrow${13.00}} & \textbf{$\uparrow${1.34}} & \textbf{$\uparrow${16.00} }& \textbf{$\uparrow${37.68}} \\
\midrule                              
\multirow{4}{*}{LLaVA 13B} & Original & $116.67_{(\pm8.50)}$ & $68.67_{(\pm7.94)}$ & $56.00_{(\pm0.91)}$ & $58.67_{(\pm3.61)}$ & 300.01\\
                               & VCD & $117.20_{(\pm21.14)}$ & $61.33_{(\pm5.45)}$ & $50.00_{(\pm3.73)}$ & $\mathbf{65.33_{(\pm3.98)}}$ & 293.86 \\
                              & PLANE & $\mathbf{125.00_{(\pm11.73)}}$ & $\mathbf{72.00_{(\pm9.82)}}$ & $\mathbf{59.67_{(\pm3.98)}}$ & $63.00_{(\pm5.70)}$ & \textbf{319.67} \\
                                 & $\Delta$ & \textbf{$\uparrow${8.33} }& \textbf{$\uparrow${3.33}} & \textbf{$\uparrow${3.67}} & \textbf{$\uparrow${4.33}} & \textbf{$\uparrow${19.66}}  \\

    \bottomrule
  \end{tabular}}
   \caption{Results on MME. The best results are shown in bold, and $\Delta$ indicates the change in performance between our method compared to the original model. Standard deviation is shown in parentheses.}
  \label{tab:MME}
\end{table*}

\subsection{Results on MME}
Table \ref{tab:MME} presents the  results on MME, showing a significant improvement in the overall score after incorporating our method compared to the original models. For each subtask, our approach effectively mitigates different types of hallucinations, which can be attributed to use of fine-grained image-caption pairs to train visual experts that capture different aspects of the information. Compared to existing hallucination mitigation methods, PLANE shows superior performance and remains effective across models of varying sizes and types.  Notably, we train PLANE only on LLaVA-v1.5 and it serves as a plug-and-play module for other LVLMs using a frozen CLIP-ViT-L encoder, effectively mitigating hallucinations across diverse tasks. As evidenced by the MME full set results in Appendix \ref{app:mmefull}, PLANE further improves overall performance beyond hallucination reduction.

\subsection{Results on GIFT and AMBER}

\begin{table}[t]
\centering
\setlength{\tabcolsep}{1pt}
\resizebox{0.48\textwidth}{!}{
\begin{tabular}{ccccccc}
\toprule
\multirow{2}{*}{Model} & \multirow{2}{*}{Method} & \multicolumn{2}{c}{GIFT} & \multicolumn{3}{c}{AMBER} \\
\cmidrule(lr){3-4} \cmidrule(lr){5-7}
& & Hallu$\downarrow$ & Cover$\uparrow$ & Hal$\downarrow$ & CHAIR$\downarrow$ & Cog$\downarrow$  \\
\midrule
\multirow{6}{*}{LLaVA-1.5}  & Original & 23.39 & 52.96 & 36.5 & 8.1 & 3.4 \\
& OHD-Caps & 22.77 & 53.06 & 36.6 & 7.8 & 3.5 \\
& HA-DPO & 23.43 & 52.78 & 35.0 & 7.7 & 3.4 \\
& VCD & 25.25 & 53.55 & 38.1 & 8.4 & 4.2 \\
& VAF & 23.05 & 53.74 & 37.9 & 8.0 & 3.9 \\
& PLANE & \textbf{20.30} & \textbf{54.07} &\textbf{33.5} & \textbf{7.5} & \textbf{3.3} \\
\midrule
\multirow{3}{*}{Mini-Gemini}  & Original & 18.64 & 83.81 & 54.3 & 9.7 & 4.9 \\
& VCD & 22.12 & 83.86 & 58.0 & 10.7 & 4.9 \\
& PLANE & \textbf{17.27} & \textbf{83.86} & \textbf{51.2} & \textbf{8.8} & \textbf{4.0} \\
\midrule
\multirow{3}{*}{Otter}  & Original & 21.81 & 25.56 & 19.8 & 8.9 & 1.5 \\
& VCD & 26.97 & 25.79 & 24.7 & 11.3 & 1.7 \\
& PLANE & \textbf{20.18} & \textbf{28.24} & \textbf{18.5} & \textbf{7.8} & \textbf{1.3} \\
\midrule
\multirow{3}{*}{Vary Toy}  & Original & 29.71 & 29.79 & 39.9 & 10.0 & 4.5 \\
& VCD & 28.67 & \textbf{31.69} & 40.4 & 10.2 & 4.1 \\
& PLANE & \textbf{28.31} & 30.12 & \textbf{37.8} & \textbf{9.6} & \textbf{4.0} \\
\bottomrule
\end{tabular}}
\caption{Results on GIFT and AMBER.}
\label{tab:GIFT}
\end{table}

The evaluation results on the GIFT and AMBER benchmarks are shown in Table \ref{tab:GIFT}. After integrating PLANE with different LVLMs, we observe a decrease in the Hallu Rate and an increase in the Cover Rate on GIFT, indicating that our method reduces fine-grained hallucinations without sacrificing the level of detail in the generated content. This enhancement is attributed to the three LoRA visual experts, each focusing on distinct visual aspects. Moreover, PLANE achieves a plug-and-play effect in generation tasks—trained only once on LLaVA1.5, it can effectively mitigate hallucinations when incorporated into other models. Experiments on AMBER further demonstrate that our approach achieves superior performance across various hallucination metrics compared to existing training-based and training-free methods. Overall, PLANE effectively reduces hallucinations across a variety of model architectures in generation tasks, demonstrating superior performance in alleviating fine-grained hallucinations compared to other training and training-free methods, which further corroborates the findings on GIFT.

\subsection{Further Analysis}

\subsubsection{Inference Speed}
\begin{figure}[t]
    \begin{center}
        \includegraphics[width=1.0\columnwidth]{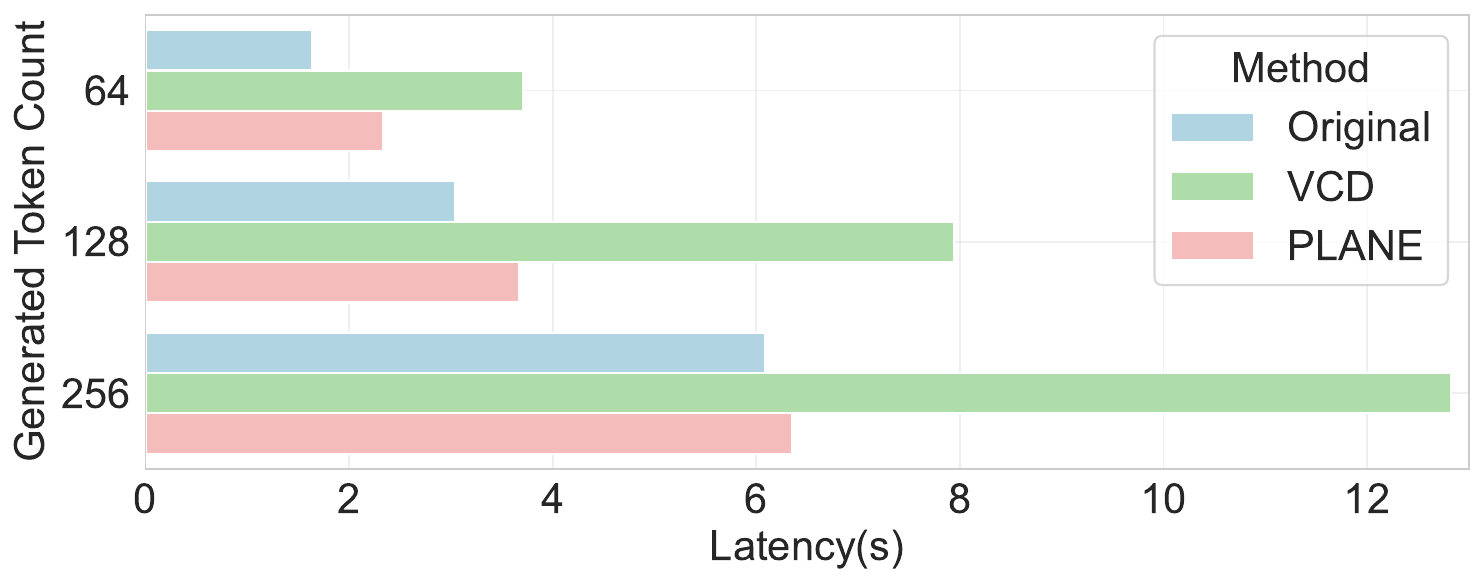}
    \end{center}
    \caption{Inference time comparison.}
    \label{fig:time}
\end{figure}

We compare inference time on LLaVA1.5, with results shown in Figure \ref{fig:time}. Our method's inference time is close to the original model when generating content of different lengths, introducing minimal latency. VCD requires nearly twice the inference time due to its contrastive decoding. This demonstrates our model's superiority in effectively mitigating hallucinations with almost no additional inference overhead.

\subsubsection{Ablation Studies}

We conduct ablation studies on MME using Vary Toy to evaluate each component's effectiveness. We test three expert encoders (object, attribute, and relation) by incorporating them individually into LVLMs to assess their impact on reducing specific types of multimodal hallucinations. We verify the effect of the Ins-Gate by removing it and taking the average of the image representations as the final image embedding. As shown in Table \ref{tab:Ablation Studies}, each specialized encoder improves performance on its corresponding subtask, while also enhancing overall model performance. While direct feature averaging shows notable improvements, optimal results are achieved with the Ins-Gate integration.

\subsubsection{Human Evaluation on GIFT}
To validate the reliability of GIFT in evaluating fine-grained hallucinations, we conduct experiments on Vary Toy using greedy decoding. We employ both LLaMA-Judge, GPT4 and human evaluators to determine whether the generated triplets contain hallucinations, measuring both Hallu Rate and Cover Rate (in Table \ref{tab:human}). We also report the consistency of LLaMA-Judge and GPT4 with human evaluations, calculated as the proportion of triplets where judgment results align with human annotations. Our trained LLaMA-Judge achieves 83.70\% consistency with human evaluators when assessing hallucinations in triplets, significantly outperforming GPT4's 71.81\%, which further confirms GIFT's reliability in assessing multimodal hallucinations. 
It's also worth noting that regardless of which model is used as the judge, PLANE consistently reduces hallucination rates and improves coverage, demonstrating PLANE's excellent performance in mitigating hallucinations.

\begin{table}[t!]
  \centering

  \setlength{\tabcolsep}{1pt}
\resizebox{0.48\textwidth}{!}{
  \begin{tabular}{cccccc}
    \toprule
    Method & Existence$\uparrow$& Count$\uparrow$& Position$\uparrow$& Color$\uparrow$ & Total$\uparrow$  \\
    \midrule
    Original & 176.33 & 86.00 & 84.33 & 87.00 & 433.66 \\
    w Object & \textbf{186.00} & 81.67 & 84.33 & 94.67 & 446.67 \\
    w  Relation & 179.00 & 97.67 & \textbf{88.67} & 102.67 & 468.01 \\
    w Attribute & 175.33 & 92.00 & 76.67 & \textbf{108.33} & 452.33\\
    w/o Ins-Gate & 180.66 & 89.00 & 87.33 & 106.00 & 462.99\\
    PLANE & 183.67 & \textbf{99.00} &85.67 & 103.00 & \textbf{471.34}\\
    \bottomrule
  \end{tabular}}
    \caption{Ablation study.}
  \label{tab:Ablation Studies}
\end{table}

\begin{table}[thb!]
\centering
\setlength{\tabcolsep}{0.75pt}
\resizebox{0.49\textwidth}{!}{
\begin{tabular}{ccccccccc}
\toprule
 & \multicolumn{2}{c}{Human} & \multicolumn{3}{c}{LLaMA-Judge} & \multicolumn{3}{c}{GPT4} \\
\cmidrule(lr){2-3} \cmidrule(lr){4-6} \cmidrule(lr){7-9} 
Method & Hallu$\downarrow$ & Cover$\uparrow$ & Hallu$\downarrow$ & Cover$\uparrow$ & CWH$\uparrow$ & Hallu$\downarrow$ & Cover$\uparrow$ & CWH$\uparrow$  \\
\midrule
Original & 33.99  & 28.04 & 32.78 & 28.56  & \multirow{2}{*}{\textbf{83.70}} & 54.82 & 19.20 & \multirow{2}{*}{71.81} \\
PLANE & \textbf{30.66} & \textbf{28.09} & \textbf{28.97} & \textbf{28.84} & & \textbf{50.41}  & \textbf{20.14} & \\
\bottomrule
\end{tabular}}
\caption{Human Evaluation on GIFT. CWH (Consistency with Human) measures the percentage of triplets where model judgments match human annotations.}
\label{tab:human}

\end{table}

\section{Conclusion}
This paper proposes a plug-and-play fine-grained vision encoder to mitigate multimodal hallucination, consisting of fine-grained LoRA visual experts and an instruction-aware gate network. The former encodes images at object-aspect, attribute-aspect, and relation-aspect respectively, while the latter fuses multiple image features based on user instructions. We also construct a fine-grained benchmark which is GPT4-
independent during evaluation phase for LVLMs. The experimental results on popular benchmarks and GIFT demonstrate the effectiveness of our proposed method. 
Additionally, Appendix evaluation shows most mainstream LVLMs exhibit high hallucination and low coverage rates on GIFT, demonstrating it as a challenging benchmark that can advance the LVLMs community.

\section*{Limitations}
The proposed fine-grained open-ended hallucination benchmark (GIFT) remains limited in scale. Given that fine-grained image annotation is highly labor-intensive and time-consuming, scaling up this benchmark is designated as a key objective for future work.

Current mitigation methods primarily focus on single-image inputs. However, multi-image scenarios \cite{mimic,mihbench} are also prevalent in real-world applications; therefore, future work will focus on mitigation strategies for multi-image inputs.
Beyond that, we will also investigate hallucination in a wider range of application scenarios \cite{medhallu,Showmaker}.

\section*{Acknowledgments}

This research is funded by the National Nature Science Foundation of China (No.62477010, No.62577022 and No.62307028), Shanghai General AI Foundation Models Program (No.2025SHZDZX026D24) and Shanghai Science and Technology Innovation Action Plan (No.24YF2710100).

\bibliography{custom}

\clearpage

\appendix

\section{Appendix}
\label{sec:appendix}

\subsection{Fine-Grained Image-Caption Training Dataset}
\label{app:1}
The statistical information of the fine-grained image-caption pairs we constructed  is shown in \ref{tab:Statistical Information}.

\begin{table}[thb!]
  \centering
  \renewcommand{\arraystretch}{1.25}
  \begin{tabular}{cc}
    \toprule
    Aspect & Count  \\
    \midrule
    Image-Object Caption Pairs & 7800  \\
    Image-Attribute Caption Pairs & 18121  \\
    Image-Relation Caption Pairs & 12454 \\
    Total & 38375 \\
    \bottomrule
  \end{tabular}
  \caption{Fine-grained image-caption dataset statistical information.}
  \label{tab:Statistical Information}
\end{table}

\subsection{Details about benchmarks}
\label{app:benchmark}
We conduct a comprehensive evaluation of our approach on both discriminative and generative tasks. In this section, we will introduce the datasets, metrics, training details, and baselines used in our experiments.

\paragraph{MME \cite{MME}} is a comprehensive multimodal LVLMs benchmark consisting of 14 sub-tasks which are discriminative task,  the LVLMs need to respond with either \textquotesingle yes\textquotesingle~or \textquotesingle no\textquotesingle. Following previous work \cite{woodpecker,vcd}, we use a subset of these tasks, including \textit{existence}, \textit{count}, \textit{color}, and \textit{position} subsets, as the hallucination evaluation set, referred to as MME-Hallu. We evaluate the hallucination mitigation performance of our model on MME-Hallu, as well as its overall capabilities on the MME-Full set. In line with \cite{MME}, we employ the sum of accuracy and accuracy+ as the evaluation metrics.

\paragraph{POPE \cite{pope}} is a VQA benchmark for evaluating object hallucination.  The benchmark incorporates three sampling strategies of varying difficulty levels. The first is the random strategy, which involves randomly sampling nonexistent objects. The second is the popular strategy, focusing on selecting frequently occurring objects. The third is the adversarial strategy, which entails choosing objects that frequently co-occur with those present in the image. We report Accuracy, Precision, Recall, F1 Score as the evaluation metrics.

\paragraph{AMBER \cite{Amber}} includes both discriminative and generative tasks. We select the generative task for evaluation, in which the LVLMs is required to generate descriptions for images, and evaluated using three metrics: \textit{CHAIR}, \textit{Hal}, and \textit{Cog}. \textit{CHAIR} measures frequency of non-existent  objects in responses, \textit{Hal} calculates the proportion of responses with hallucinations, \textit{Cog} compares model hallucinations to human cognitive patterns.

\paragraph{GIFT} is a GPT4-independent fine-grained benchmark hallucination evaluation introduced in this paper.

\subsection{Details about baselines}
We conduct comprehensive comparisons between our proposed method and current state-of-the-art approaches in both categories: training-free models and those requiring fine-tuning. Among training-free methods, we benchmark against VCD \cite{vcd}, DROPOUT DECODING \cite{dropout} and VAF \cite{VAF}. For approaches requiring training, we compare with OHD-Caps \cite{OHD}, HADPO \cite{hadpo}, and HALVA \cite{halva} to thoroughly evaluate our method's effectiveness. 

\begin{itemize}[leftmargin=*, align=left]
\item VCD \cite{vcd} is a contrastive decoding method that mitigates hallucinations by determining the final output distribution through the combined outputs of both the perturbed image and the original image.

 \item DROPOUT DECODING \cite{dropout} measures uncertainty in visual inputs and selectively dropping out visual tokens where uncertainty is high.

 \item VAF \cite{VAF} reduces hallucinations by amplifying visual features during modality fusion, decreasing reliance on language priors while maintaining inference speed and generation quality, ensuring better consistency between visual inputs and generated content.

\item OHD-Caps \cite{OHD} uses the OHD-Caps training set to train a CLIP model with reduced hallucination as the vision encoder for LVLMs. While this method was not given a specific name in the paper, we refer to it as OHD-Caps in this work.

\item HA-DPO \cite{hadpo} mitigates hallucinations by training LVLMs to prefer non-hallucinatory outputs through direct preference optimization.

\item HALVA \cite{halva} uses data-augmented phrase-level alignment loss to train LVLMs to reduce the likelihood of hallucinated phrases compared to the correct ones.

\end{itemize}

\subsection{Details about GIFT}
\label{app:2}

\begin{table*}[thb!]

  \centering
  \resizebox{\textwidth}{!}{
  \begin{tabular}{ccccccc}
    \toprule
    Dataset& Setting & Method & Accuracy $\uparrow$ & Precision & Recall & F1 Score $\uparrow$ \\
    \midrule
\multirow{15}{*}{A-OKVQA} & \multirow{4}{*}{Random} & Regular & 83.45 &87.24 &78.36 &82.56\\
                              &                    & VCD \cite{vcd} & 86.15 &85.18 &87.53 &86.34 \\
                               &                    & RITUAL \cite{ritual} &85.17 &79.79&- &86.40 \\
                               &                    & DeGF \cite{DeGF} &\underline{86.93} &84.28&- &\underline{87.42} \\
                             &                    &  Ours & \textbf{88.93}  &95.13	&82.07	&\textbf{88.11} \\
\cmidrule{2-7}
                              & \multirow{4}{*}{Popular} & Regular & 79.90 &80.85 &78.36 &79.59\\
                              &                    & VCD \cite{vcd}& \underline{81.85} &78.60 &87.53 &\underline{82.82}\\
                               &                    & RITUAL \cite{ritual} &78.83 &71.99&- &81.68 \\
                               &                    & DeGF \cite{DeGF}  &80.90 &75.68&- &82.66  \\
                             &                    &  Ours & \textbf{85.3}	&87.79	&82.00	&\textbf{84.79} \\ 

\cmidrule{2-7}
                              & \multirow{4}{*}{Adversarial} & Regular & 74.04 &72.08 &78.49 &75.15\\
                              &                    & VCD \cite{vcd}& \underline{74.97} &70.01 &87.36 & \underline{77.73}\\
                               &                    & RITUAL \cite{ritual} &68.57 &62.26&- &74.99 \\
                               &                    & DeGF \cite{DeGF} &72.70 &66.70 &-&76.86 \\
                             &                    &  Ours & \textbf{78.63}	&76.79	&82.07	&\textbf{79.34 }\\ 
\midrule
\multirow{15}{*}{GQA}  & \multirow{4}{*}{Random} & Regular & 83.73 &87.16 &79.12 &82.95 \\
                              &                    & VCD \cite{vcd}& 86.65 &84.85 &89.24 &86.99 \\
                               &                    & RITUAL \cite{ritual} &86.10 &80.30&- &\underline{87.31} \\
                               &                    & DeGF \cite{DeGF} &\underline{87.40} &83.51 &- &\textbf{88.09} \\
                             &                    &  Ours & \textbf{87.93} &94.31	&80.73	&87.00 \\
\cmidrule{2-7}
                              & \multirow{4}{*}{Popular} & Regular & 78.17 &77.64 &79.12&78.37 \\
                              &                    & VCD \cite{vcd}& \underline{80.73} &76.26 &89.24 &\textbf{82.24}\\
                               &                    & RITUAL \cite{ritual} &74.80 &67.50&-& 79.15 \\
                               &                    & DeGF \cite{DeGF} &78.10 &71.56 &- &80.98  \\
                             &                    &  Ours & \textbf{81.80}	&82.63	&80.53	&\underline{81.57} \\ 

\cmidrule{2-7}
                              & \multirow{4}{*}{Adversarial} & Regular & 75.08 &73.19 &79.16 &76.06 \\
                              &                    & VCD \cite{vcd}& \underline{76.09} &70.83 &88.75 &\underline{78.78}\\
                               &                    & RITUAL \cite{ritual} &68.23 &61.75 &-&75.10\\
                               &                    & DeGF \cite{DeGF} &74.07 &67.42&- &78.22  \\
                             &                    &  Ours &\textbf{79.43}	&78.72	&80.67	&\textbf{79.68} \\ 
    \bottomrule
  \end{tabular}}
  \caption{Results on POPE. Best results are in bold, and second-best values are underlined.}
  \label{tab:POPE}
\end{table*}

\subsubsection{Evaluating LVLMs on GIFT}
We evaluate mainstream LVLMs on the GIFT benchmark in Table \ref{tab:mllm}. Our experimental results show that most models exhibit high hallucination rates and low coverage scores. Even the state-of-the-art GPT-4V demonstrates significant hallucination tendencies. These findings suggest that GIFT serves as a challenging hallucination evaluation benchmark, providing valuable insights and directions for the future development of the LVLMs research community.

\begin{table}[h!]
\centering
\setlength{\tabcolsep}{3pt}
\resizebox{0.47\textwidth}{!}{
\begin{tabular}{ccccc}
\toprule
Model  & Hallu Rate$\downarrow$ & Cover Rate$\uparrow$ \\
\midrule
\multirow{1}{*}{Vary Toy 1.8B}   & 29.7 & 29.8 \\
\multirow{1}{*}{LLaVA 7B}   & 19.0 & 53.2 \\
\multirow{1}{*}{LLaVA 13B}  & 19.4 & 53.4 \\
\multirow{1}{*}{LLaVA-1.5 7B}  & 23.4 & 53.0 \\
\multirow{1}{*}{QwenVL 7B} & 14.3 & 21.4 \\
\multirow{1}{*}{Cambrian 8B}  & 19.3 & 66.0 \\
\multirow{1}{*}{LLaVA-OneVision 7B}  & 17.7 & 72.3 \\
\multirow{1}{*}{GPT4V}  & 20.9 & 81.1 \\
\bottomrule
\end{tabular}}
\caption{Results on GIFT. The best results are shown in bold.}
\label{tab:mllm}

\end{table}

\subsubsection{Comparison with Existing Benchmarks}
Table \ref{tab:benchmark} shows the comparison between our proposed GIFT and existing generative hallucination evaluation benchmarks. We compare four dimensions: whether object-level, attribute-level, and relation-level hallucinations are considered, and whether ChatGPT or GPT-4 assistance is required for judgment. We can see that CHAIR \cite{chair}, AMBER \cite{Amber} (here we only discuss the generation task in AMBER), and GAVIE \cite{LVR} mainly evaluate object-level hallucinations. Faithscore \cite{faithscore} can evaluate hallucinations at multiple levels but requires ChatGPT assistance for judgment, which increases costs and affects its scalability. Our GIFT can detect fine-grained hallucinations while not requiring assistance from ChatGPT/GPT-4 during the evaluation phase. 
\begin{table}[h!]
  \centering
  \setlength{\tabcolsep}{2pt} 
\resizebox{0.47\textwidth}{!}{
  \begin{tabular}{ccccc}
    \toprule
    Benchmark & Object & Attribute & Relation & GPT Free  \\
    \midrule
    CHAIR  & \ding{51}& \ding{55}  & {\ding{55}} & {\ding{51}}  \\
    AMBER  & {\ding{51}} & {\ding{55}} & {\ding{55}} & {\ding{51}}  \\
    GAVIE  & {\ding{51}} & {\ding{55}} & {\ding{55}} & {\ding{55}}  \\
    Faithscore  & {\ding{51}} & {\ding{51}} & {\ding{51}} & {\ding{55}} \\
    GIFT & {\ding{51}} & {\ding{51}} & {\ding{51}} & {\ding{51}} \\
    \bottomrule
  \end{tabular}}
    \caption{Comparison between GIFT and existing generative hallucination benchmarks. Object, Attribute, and Relation denote coverage of corresponding hallucination types, while GPT-free indicates whether the evaluation runs without relying on ChatGPT/GPT-4.}
  \label{tab:benchmark}
\end{table}

\subsubsection{Prompt for LLaMA-Judge}
\begin{figure*}[]
  \centering
  \includegraphics[width=1\textwidth]{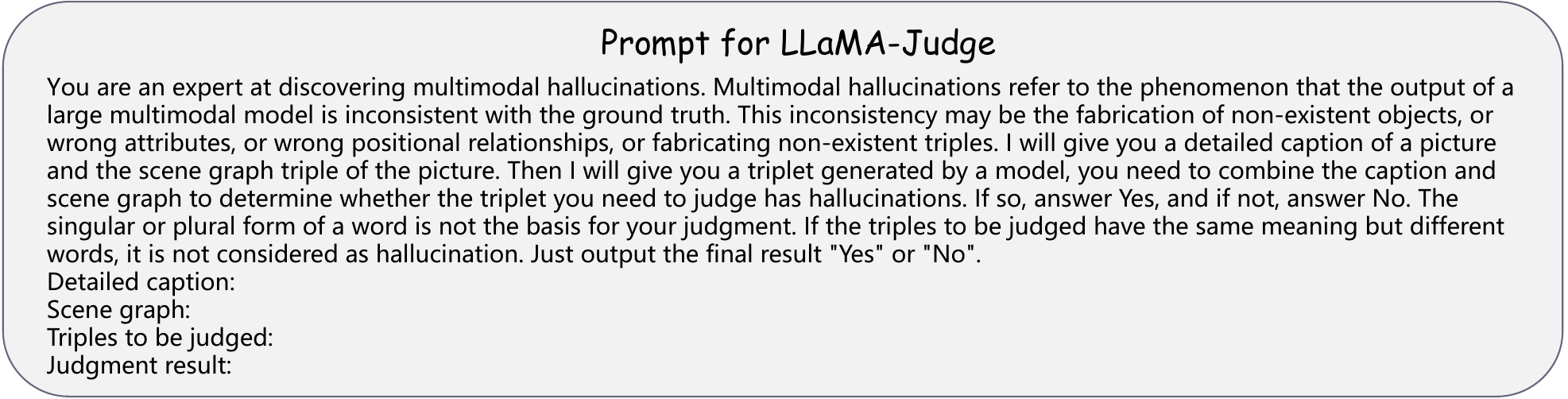}\\
  \caption{Prompt for LLaMA-Judge.}
  \label{prompt}
\end{figure*}

The prompt for LLaMA-Judge is shown in Figure \ref{prompt}, where the detailed caption in DCI \cite{denseimage} and manually annotated scene graph triples are concatenated to \textquotesingle Detailed caption\textquotesingle~ and \textquotesingle Scene graph\textquotesingle~ respectively. The triples extracted from LVLMs output are concatenated after \textquotesingle Triples to be judged\textquotesingle~.

\subsubsection{LLaMA-Judge Training Details}
To train a model that can detect hallucinations in extracted triples without relying on ChatGPT/GPT-4, we train Llama-Judge based on Llama 3 \cite{llama3}. First, we construct training data by instructing GPT-4o to generate triples both with and without hallucinations. We use uniform sampling with a positive-to-negative sample ratio close to 1:1. For training, we use LoRA fine-tuning with the rank of 8 and a learning rate of $5.0 \times 10^{-5}$.

\subsection{Results on MME Full Set}
\label{app:mmefull}
\begin{figure*}[]
  \centering
  \includegraphics[width=1\textwidth]{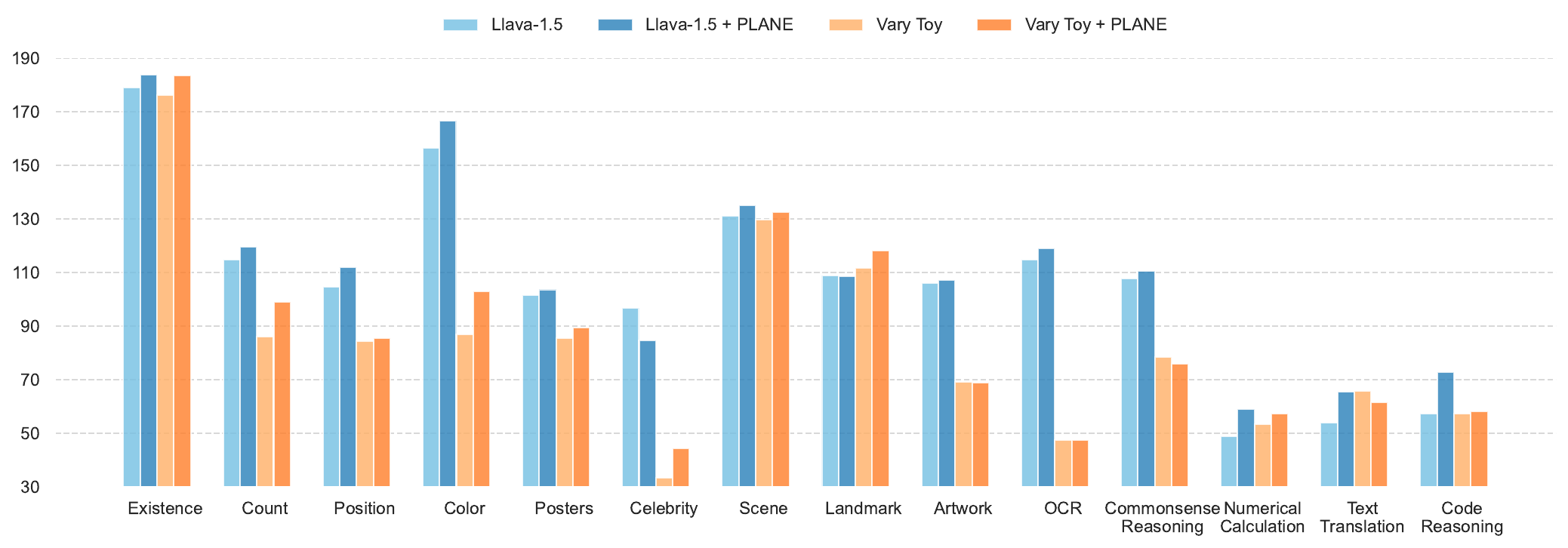}\\
  \caption{Results on MME-Full set.}
  \label{fig:MME-Full}
\end{figure*}

We conduct experiments on the MME-Full set using LLaVA-1.5 and Vary Toy, with results shown in Figure \ref{fig:MME-Full}. After incorporating PLANE, different models show varying trends across different subtasks, which may be attributed to differences in model architectures and training data. However, the implementation of PLANE leads to performance enhancements across the majority of subtasks compared to the baseline. Overall, our model not only excels at hallucination mitigation but also enhances performance on general tasks compared to the original model.

\subsection{Results on POPE}
\label{app:pope}
We evaluate our method on POPE using LLaVA v1.5, and the experimental results are presented in Table \ref{tab:POPE}. Our method yields improvements under the three settings of Random, Popular, and Adversarial. Compared with existing methods, it achieves the best Accuracy and, in most cases, the best F1 score as well.

\subsection{Parameters Efficiency Analysis}
In this section, we compare the trainable parameter counts with methods that require training. The experimental results are shown in Table \ref{tab:Trainable Parameters}. PLANE has fewer parameters than the other two training-required methods, HA-DPO and HALVA, while achieving better performance (Table \ref{tab:MME}).

\begin{table}[thb!]
  \centering
  \renewcommand{\arraystretch}{1.25}
  \begin{tabular}{cc}
    \toprule
    Method & Trainable Parameters \\
    \midrule
    HA-DPO & 340.8 M  \\
    HALVA & 340.8 M  \\
    PLANE & \textbf{332.4 M} \\
    \bottomrule
  \end{tabular}
  \caption{Trainable Parameters Comparison.}
  \label{tab:Trainable Parameters}
\end{table}

\begin{table}[]
  \centering
  \setlength{\tabcolsep}{1pt}
\resizebox{0.48\textwidth}{!}{
  \begin{tabular}{cccccc}
    \toprule
    Method & Existence$\uparrow$& Count$\uparrow$& Position$\uparrow$& Color$\uparrow$ & Total$\uparrow$  \\
    \midrule
    Original & 176.33 & 86.00 & 84.33 & 87.00 & 433.66 \\
    All in one 	&165.00	&90.00	&86.67	&\textbf{103.89}	&445.56\\
    PLANE & \textbf{183.67} & \textbf{99.00} &\textbf{85.67} & 103.00 & \textbf{471.34}\\
    \bottomrule
  \end{tabular}}
    \caption{Ablation on the necessary of Separate Experts.}
  \label{tab:Necessary}
\end{table}

\subsection{Response Length Analysis}

To demonstrate that the hallucination reduction on GIFT does not arise from shortening the model's output, we evaluate the output length at the triplet level. We conduct experiments on LLaVA-1.5 and Vary Toy, using GIFT's scene graph parser to extract the triplets from the model outputs. The experimental results are shown in Table \ref{tab:length}. We can observe that PLANE barely changes the model's output length, with only negligible differences from the original model. This confirms that the reduction in hallucinations does not come from shortening the model's output length.

\begin{table*}[]
  \centering
  \renewcommand{\arraystretch}{1.25}
  \setlength{\tabcolsep}{3pt} 
  \begin{tabular}{ccccc}
    \toprule
    \textbf{Model} & \textbf{Method} & \textbf{Length (Avg. Triplets)} & \textbf{Hallu Rate (\%)} & \textbf{Cover Rate (\%)} \\
    \midrule
    \multirow{2}{*}{LLaVA-1.5} & Original & 7.18 & 23.39 & 52.96 \\
    & PLANE & 7.03 & \textbf{20.30} & \textbf{54.07} \\
    \midrule
    \multirow{2}{*}{Vary-Toy} & Original & 5.02 & 29.71 & 29.79 \\
    & PLANE & 5.08 & \textbf{28.31} & \textbf{30.12} \\
    \bottomrule
  \end{tabular}
  \caption{Response Length Analysis.}
  \label{tab:length}
\end{table*}

\subsection{Necessary of Separate Experts}

To verify the necessity of the separate experts design in PLANE, we combine all training tree caption types into a single expert (denoted as "All in one"). The experimental results are shown in Table \ref{tab:Necessary}. "All in one" underperforms compared to PLANE across most subtasks, which demonstrates the necessity of the separate experts design.

\subsection{Extended Ablation Study}
\label{app:3}

In this section, we will explore how different fusion strategies in Ins-Gate affect the results. We compare two fusion methods: one is to fuse the image feature as a whole feature, where \(gate \in \mathbb{R}^{1\times 4}\), which we denote as Image-Level; the other is to fuse the image feature by token, where \(gate \in \mathbb{R}^{n\times 4}\) (\(n\) represents the number of tokens contained in the image feature), which we denote as Token-Level. We conduct experiments using LLaVA-1.5 and Vary Toy on MME-Hallu, and the experimental results are shown in Table~\ref{tab:image-level}. We find that both fusion strategies can mitigate hallucinations compared to the original model, with Token-Level achieving better results.

\begin{figure*}[]
  \centering
  \includegraphics[width=1\textwidth]{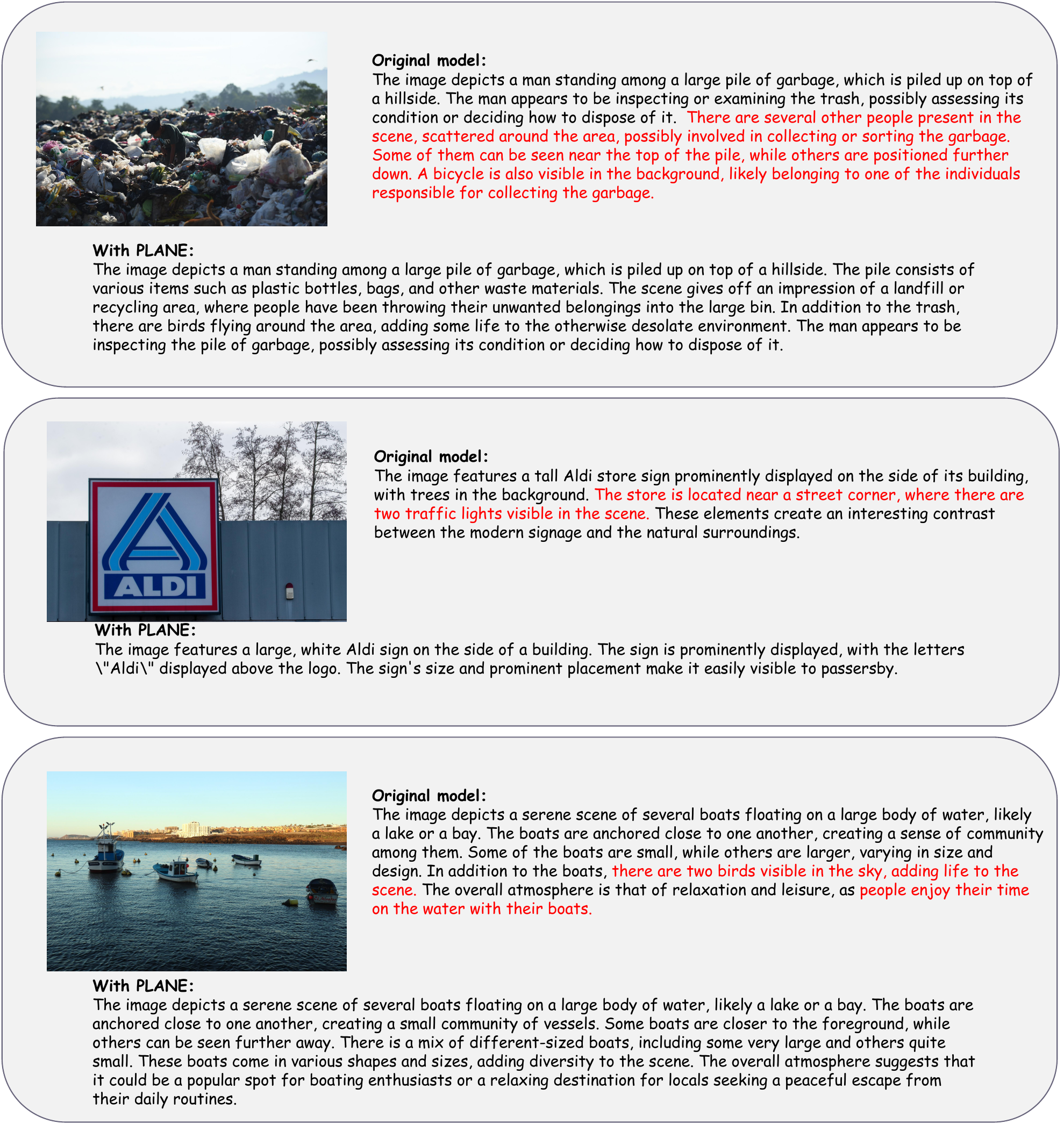}\\
  \caption{Case study on GIFT. The sentences that experience hallucinations are marked in red.}
  \label{case}
\end{figure*}

\begin{table*}[h!]
\small
\setlength{\tabcolsep}{2pt}
  \centering

  \begin{tabular}{ccccccc}
    \toprule
    Model & Method & Existence$\uparrow$ & Count $\uparrow$& Position$\uparrow$& Color$\uparrow$ & Total$\uparrow$  \\
    \midrule
\multirow{3}{*}{LLaVA-1.5 7B}  & Original & $179.00_{(\pm2.24)}$ & $115.00_{(\pm15.76)}$ & $104.67_{(\pm20.26)}$ & $156.71_{(\pm4.88)}$ & 555.38 \\
                              &Image-Level & $\mathbf{184.00_{(\pm2.24)}}$ & $113.34_{(\pm1.67)}$ & $\mathbf{112.33_{(\pm16.01)}}$ & $151.33_{(\pm15.20)}$ & 561.00 \\
                             &Token-Level& $\mathbf{184.00_{(\pm4.18)}}$  & $\mathbf{119.67_{(\pm5.05)}}$ & $112.00_{(\pm13.10)}$ & $\mathbf{166.67_{(\pm12.19)}}$ & \textbf{582.34} \\
\midrule                              
\multirow{3}{*}{Vary Toy 1.8B} & Original & $176.33_{(\pm2.74)}$ & $86.00_{(\pm2.74)}$ & $84.33_{(\pm2.74)}$ & $87.00_{(\pm9.76)}$  & 433.66 \\
                              & Image-Level & $174.67_{(\pm3.41)}$ & $90.00_{(\pm5.27)}$ & $\mathbf{90.67_{(\pm5.22)}}$ & $89.00_{(\pm2.24)}$  & 444.34 \\
                              & Token-Level & $\mathbf{183.67_{(\pm3.20)}}$ & $\mathbf{99.00_{(\pm2.74)}}$ & $85.67_{(\pm2.74)}$ & $\mathbf{103.00_{(\pm1.90)}}$  & \textbf{471.34} \\
    \bottomrule
  \end{tabular}
      \caption{Comparison of different fusion strategies. The best results are shown in bold. Standard deviation is shown in parentheses.}
  \label{tab:image-level}
\end{table*}

\subsection{Case Study}
\label{app:4}
To more intuitively demonstrate the effects of PLANE, we conducted a case study on GIFT. Figure \ref{case} shows three cases, displaying the output of Vary Toy and the output after incorporating PLANE. And the sentences where the hallucinations occurred are marked in red. We can observe that the original model easily outputs object-level hallucinations and relation-level hallucinations. After incorporating PLANE, there is a great degree of mitigation of fine-grained hallucinations, while also preserving the semantic integrity and coherence. These cases also verify the advantages of PLANE in mitigating hallucinations.

\end{document}